\documentclass[runningheads]{llncs}
\usepackage{graphicx}
\usepackage[utf8]{inputenc}
\usepackage[T1]{fontenc}
\usepackage[british]{babel}
\usepackage{csquotes}
\usepackage{amsmath}
\usepackage{amssymb}
\usepackage{mathtools}
\usepackage{acronym}
\usepackage[dvipsnames]{xcolor}
\usepackage{tabularx}
\usepackage{xspace}
\usepackage[caption=false]{subfig}
\usepackage{tikz}
\usepackage{fixme}
\usepackage{prftree}

\fxuselayouts{inline,nomargin}
\fxusetheme{color}
\FXRegisterAuthor{ls}{als}{LS}
\FXRegisterAuthor{mg}{amg}{MG}

\usetikzlibrary{calc, patterns, arrows, shapes, positioning, fit}

\usepackage{hyperref}

\acrodef{AIF}{Argument Interchange Format}
\acrodef{IAT}{Inference Anchoring Theory}
\acrodef{ASD}{Argumentation Scheme Dialogue}
\acrodef{S-AIF}{Social \acl*{AIF}}
\acrodef{T-AIF}{Trichotomic \acl*{AIF}}
\acrodef{C-AF}{C-Argumentation Framework}
\acrodef{T-AF}{Trichotomic Argumentation Framework}
\acrodef{BAF}{Bipolar Argumentation Framework}
\acrodef{AAF}{Abstract Argumentation Framework}
\acrodef{OWL}{Web Ontology  Language}

\DeclareMathOperator{\as}{as}   %
\DeclareMathOperator{\es}{es}   %
\DeclareMathOperator{\si}{si}   %

\usepackage[utf8]{inputenc}

\newcolumntype{Y}{>{\centering\arraybackslash}X}

\spnewtheorem{thm}[theorem]{Theorem}{\bfseries}{\itshape}
\spnewtheorem{cor}[theorem]{Corollary}{\bfseries}{\itshape}
\spnewtheorem{cnj}[theorem]{Conjecture}{\bfseries}{\itshape}
\spnewtheorem{lem}[theorem]{Lemma}{\bfseries}{\itshape}
\spnewtheorem{lemdefn}[theorem]{Lemma and Definition}{\bfseries}{\itshape}
\spnewtheorem{prop}[theorem]{Proposition}{\bfseries}{\itshape}
\spnewtheorem{defn}[theorem]{Definition}{\bfseries}{\upshape}
\spnewtheorem{rem}[theorem]{Remark}{\bfseries}{\upshape}
\spnewtheorem{notation}[theorem]{Notation}{\bfseries}{\upshape}
\spnewtheorem{expl}[theorem]{Example}{\bfseries}{\upshape}
\spnewtheorem{thmdefn}[theorem]{Theorem and Definition}{\bfseries}{\itshape}
\spnewtheorem{propdefn}[theorem]{Proposition and Definition}{\bfseries}{\itshape}
\spnewtheorem{assumption}[theorem]{Assumption}{\bfseries}{\upshape}
\spnewtheorem{algorithm}[theorem]{Algorithm}{\bfseries}{\upshape}

 \renewenvironment{definition}{\begin{defn}}{\end{defn}}
 \renewenvironment{remark}{\begin{rem}}{\end{rem}}
 \renewenvironment{example}{\begin{expl}}{\end{expl}}
 
\newcommand*{\green}[1]{\textcolor{Green}{#1}}

\makeatletter
\newif\ifdraft
\@ifclasswith{llncs}{draft}{\drafttrue}{\draftfalse}
\makeatother
\newcommand{\draftonly}[1]{\ifdraft \green{#1}\fi}

\renewcommand{\implies}{\rightarrow}
\renewcommand{\iff}{\leftrightarrow}

\newcommand*{\aifp}{AIF{\kern-.01em\raise.75ex\hbox{\scriptsize+}}\xspace}
\newcommand*{\aspic}{ASPIC{\kern-.01em\raise.75ex\hbox{\scriptsize+}}\xspace}

\newcommand*{\N}{\mathbb{N}}

\newcommand*{\defeq}{\coloneqq}

\newcommand*{\truth}{[0,1]}

\newcommand*{\Sim}{\mathit{Sim}}
\newcommand*{\Att}{\mathit{Att}}
\newcommand*{\Sup}{\mathit{Sup}}
\newcommand*{\Def}{\mathit{Def}}
\newcommand*{\Co}{\mathit{Co}}
\newcommand*{\Al}{\mathit{Adm}}
\newcommand*{\Sl}{\mathit{Sl}}
\newcommand*{\Pl}{\mathit{Pref}}
\newcommand*{\Cl}{\mathit{Cl}}
\newcommand*{\Gl}{\mathit{Gl}}

\newcommand*{\Ag}{\mathit{Ag}}
\newcommand*{\Ra}{\mathit{Ra}}
\newcommand*{\Jt}{\mathit{Jt}}
\newcommand*{\Tc}{\mathit{Tc}}

\newcommand*{\aas}{argumentation action\xspace}
\newcommand*{\aap}{argumentation actions\xspace}

\begin{document}
\title{Trichotomic Argumentation Representation\thanks{Work forms part of the DFG Project ``Reconstructing Arguments from Noisy Text'' (RANT) within SPP 1999 RATIO.}}
\titlerunning{Trichotomic Argumentation Representation}
\author{Merlin Göttlinger\orcidID{0000-0002-2251-8519} \and Lutz Schröder\orcidID{0000-0002-3146-5906}}
\authorrunning{M. Göttlinger \and L. Schröder}
\institute{Friedrich-Alexander-Universität Erlangen-Nürnberg}
\maketitle              %
\begin{abstract}
  The Aristotelian trichotomy distinguishes three aspects of
  argumentation: Logos, Ethos, and Pathos.  Even rich argumentation
  representations like the \ac*{AIF} are only concerned with capturing
  the Logos aspect.  \ac*{IAT} adds the possibility to represent
  ethical requirements on the illocutionary force edges linking
  locutions to illocutions, thereby allowing to capture some aspects
  of ethos. With the recent extensions \aifp and \ac*{S-AIF}, which
  embed dialogue and speakers into the \acs*{AIF} argumentation
  representation, the basis for representing all three aspects
  identified by Aristotle was formed. In the present work, we develop
  the \ac*{T-AIF}, building on the idea from \acs*{S-AIF} of adding
  the speakers to the argumentation graph. We capture Logos in the
  usual known from \aifp, Ethos in form of weighted edges between
  actors representing trust, and Pathos via weighted edges from actors
  to illocutions representing their level of commitment to the
  propositions.  This extended structured argumentation representation
  opens up new possibilities of defining semantic properties on this
  rich graph in order to characterize and profile the reasoning
  patterns of the participating actors.

\keywords{Logos \and Ethos \and Pathos \and Structured Argumentation \and Trust \and Dialogue Structure \and Argument Interchange Format \and Inference Anchoring}
\end{abstract}
\section{Introduction}
\label{sec:introduction}

Argumentation plays a central role in society for forming rational opinions about controversial topics by providing a way to resolve conflicting views.
For understanding complex argumentative discussions it is helpful to draw diagrams called \emph{argument maps} that visualize how the propositions and arguments interact with one another.
Having such a formal representation for argumentation also allows for having computational semantics for them to aid humans in evaluating argumentation.

Over the years there have been many developments of such argumentation frameworks with varying information content and complexity.
One of the simplest forms of argumentation frameworks are the so called \acp{AAF} that treat arguments as abstract objects without having attached internal structure.

\paragraph{\aclp{AAF}} consist of a graph having arguments as nodes without any additional structure or information apart from the relations connecting them to other arguments.
Dung~\cite{Dung_1995} popularized the simplest form of these frameworks (see \autoref{def:dung_aaf}) with only a single relation representing attack between arguments.
\begin{definition}[Dung \ac{AAF} \cite{Dung_Thang_2018}]\label{def:dung_aaf}
  An \emph{abstract argumentation framework} $AF = (AR, \to)$ is composed of a set $AR$ of \emph{arguments} and an \emph{attack} relation $\to\; \subseteq AR \times AR$. An argument $A \in AR$ is called an \emph{attacker} of $B \in AR$ if $A \to B$. A set of arguments $S \subseteq AR$ \emph{attacks} $B$ if there exists $A \in S$ s.t.\ $A \to B$. $S$ \emph{defends} $A$ if $S$ attacks each attacker of $A$. $S$ is \emph{conflict-free} if it does not attack its own arguments. $S$ is \emph{admissible} if it is conflict-free and defends each of its arguments. The \emph{characteristic function} of $AF$ is defined by $F_{AF}(S) \defeq \{A \in AR \mid S \text{ defends } A\}$. $S \subseteq AR$ is:
  \begin{itemize}
    \item a \emph{stable extension} if it is conflict-free and attacks each argument $A \not\in S$;
    \item a \emph{preferred extension} if it is a maximal (w.r.t.\ set inclusion) admissible set of arguments;
    \item a \emph{complete extension} if it is admissible and contains each argument it defends (or equivalently a conflict-free fixed point of $F_{AF}$);
    \item a \emph{grounded extension} if it is the least complete extension (or equivalently the least fixed point of $F_{AF}$).
  \end{itemize}
\end{definition}
\noindent While being algorithmically and conceptionally simple, %
having only an attack relation limits the expressivity of the framework.
In argumentation, it is often the case that one argument supports the truth of another argument, and it is not always easy to express this as an attack.
\acp{BAF}~\cite{CayrolLagasquieSchiex05} add a support relation between arguments to capture those situations.
The semantics of \acp{BAF} is a priori somewhat less clear-cut than that of AAF, however, as one needs to determine how support and attach interact.

One approach, extensively discussed by Cohen et al.~\cite{Cohen_2018}, is to derive \emph{complex attacks} from joining support and attack relations and thus reduce to Dung semantics.
Another way is to move from  crisp notions of \emph{accept} and \emph{reject} towards a weighted acceptance where supporters and attackers are accumulated into the acceptability of an argument \cite{Amgoud_2017,mossakowski18:modul_seman_charac_bipol_weigh_argum_graph,amgoudweighted}.
The weighted approach however comes at the cost of requiring acyclicity of the graph as evident in the overview of current bipolar semantics by Amgoud and Ben-Naim~\cite{Amgoud_2017}.

\paragraph{Structured Argumentation Frameworks} go beyond this abstract view of argumentation and attach structure to the argument map.
There are multiple points where one can add information to obtain a richer argument representation.
One way is to add structure to the nodes, e.g.\ in the form of logical formulas representing the propositions and reasoning patterns used in an argument \cite{BesnardHunter01,Caminada_2007,AmgoudBesnard09,Modgil_2014}.
This allows using logical reasoning to find conflicts and inferences between the arguments and generating new arguments from existing facts and inference rules.
Structure can also be added to the edges in the form of identifying different kinds of inferences, conflicts, or exceptions.
Argument structure and the classification of different types of arguments have been studied widely~\cite{Toulmin58,perelman1973new,Walton95,walton_reed_macagno_2008,Wagemans_2016}.

Toulmin~\cite{Toulmin58} proposed that arguments generally obey a basic structure, commonly referred to as the \emph{Toulmin Scheme}, consisting of identifiable parts that justify why a listener should believe the qualified conclusion.
Perelman and Olbrechts-Tyteca~\cite{perelman1973new} informally categorized different types of arguments and analysed what makes them convincing.
Walton~\cite{Walton95} combined these efforts and created a compendium of different \emph{Argumentation Schemes} with a common structure of identifying multiple premises necessary to reasonably derive the truth of an argument's claim.
Additionally, he identified so called \emph{critical questions} specific to each scheme, which a critically thinking audience can put forth to undermine arguments employing that scheme, as in the following example.
\begin{example}[Scheme: Argument from Position to Know \cite{walton_reed_macagno_2008}]\label{ex:pos}
  \begin{description}
    \item[Major Premise:] Source $a$ is in a position to know about things in a certain subject domain $S$ containing proposition $A$
    \item[Minor Premise:] $a$ asserts that $A$ is true (false)
    \item[Conclusion:] $A$ is true (false)
    \item[Critical Questions:] \begin{enumerate}
      \item Is $a$ in a position to know whether $A$ is true (false)?
      \item Is $a$ an honest (trustworthy, reliable) source?
      \item Did $a$ assert that $A$ is true (false)?
      \end{enumerate}
  \end{description}
\end{example}
For an in-depth review of this style of argumentation schemes as well
as a historical overview of classifying different patterns of
argumentation see the book by Walton et
al.~\cite{walton_reed_macagno_2008}.  Due to their nature of clearly
identifying premises and claim of an argument, Walton-style schemes
are a good candidate for complex edge labels in an argumentation map.

\paragraph{The \ac{AIF}} is a formal representation of argumentation general enough to include many existing representations in order to establish a standard for storing and exchanging argumentation \cite{CHES_EVAR_2006}.
Inferences in \ac{AIF} can be labelled with instances of \emph{rule schemes} which can be Walton--style argumentation schemes but are not limited to those.
Some of these schemes are implemented in the \emph{\ac{AIF} ontology}, which specifies the overall format and accompanying concepts intended as labels for the participating propositions.

To represent argumentative discourse between multiple actors, the \ac{AIF} was later extended to \aifp  by Modgil and McGinnis~\cite{Modgil} and by Reed et al.~\cite{DBLP:conf/comma/ReedWDR08}, who  introduce new types of nodes representing the dialogue structure through dialogue moves relating locutions.

\ac{IAT} introduced the idea to incorporate illocutionary force into argument maps for relating locutions and illocutions as well as transitions and inferences/conflicts \cite{budzynska2011speech,DBLP:conf/comma/BudzynskaJKRSSY14,DBLP:journals/argcom/BudzynskaJRS16}. %
Having such links between statements and their content also allows specifying requirements on force edges, e.g.\ that the speaker is trustworthy.
This can be seen as capturing ethical requirements similar to premises of Walton-style argumentation schemes, expressing that a witness is trustworthy in an \emph{Argument from Witness Testimony}.
In the extension by Reed et al.~\cite{DBLP:conf/comma/ReedWBD10}, \aifp already has the capabilities necessary to represent \ac{IAT} annotated argumentation maps.

Recently, Lawrence et al.~\cite{Lawrence_2017} formalized an adjunct
ontology to \aifp called the \emph{\ac{S-AIF}}, which incorporates the
authors of locutions into the model.  They use this to calculate
various statistics about the discussion, such as participation and
agreement-based author clustering.  We build on this idea in
Section~\ref{sec:taif} where we propose the \emph{\acl*{T-AIF}}
\acs*{T-AIF}, which formally captures even more aspects of
argumentation.  \ac{S-AIF} distinguishes between the roles of Argument
Web participants and is tailored towards augmented argument
construction and statistics, while our aim is to represent existing
arguments in more detail.

Argumentation in \aifp is represented through different types of connected nodes with corresponding semantic meaning.
The nodes in the actual argumentation graph are instances of the concepts from the \emph{upper ontology} with related instances of concepts from the \emph{forms ontology} specifying the schemes for the S-nodes as well as concepts to classify the I-nodes into the various forms (see the appendix).

\emph{I-nodes} represent \textbf{I}llocutions and hence contain the
propositions relevant for the arguments as premises, exceptions, and conclusions.  The \ac{AIF} %
defines concepts necessary
to classify these nodes to participate in Walton-style argumentation
schemes.

\emph{L-nodes} represent \textbf{L}ocutions and contain the raw utterances that make up the represented dialogue.
They can be seen as  illocutions in the sense that they carry the propositional content that their text was uttered in the conversation.
In \ac{S-AIF}, L-nodes are connected with nodes representing the authors of the locution.

\emph{S-nodes} represent \textbf{S}chemes for interconnecting nodes.
These schemes are associated with scheme application nodes for the various kinds of interconnections between nodes:
\begin{description}
  \item[RA] Instances of rule schemes, such as Walton-style schemes.
  \item[TA] Instances of transition schemes representing speech acts in dialogue.
  \item[CA] Instances of conflict schemes capture that truth of propositions is in conflict.
  \item[PA] Instances of preference schemes, which resolve conflicts via preference rules.
  \item[YA] Instances of illocutionary schemes modelling illocutionary force as a link between locutions, illocutions, inferences, and transitions.
\end{description}
The semantics of \ac{AIF} graphs is defined by translating into
ASPIC~\cite{DBLP:conf/comma/BexPR10} and then evaluating using, e.g.,
the TOAST algorithm \cite{DBLP:conf/comma/SnaithR12}.

\paragraph{ASPIC} is a framework for building structured argumentation
systems with a strong logical background following the ideas of
Besnard and Hunter~\cite{BesnardHunter01} where the nodes are sets of
formulas representing a derivation tree
\cite{Prakken_2010,DBLP:conf/comma/BexPR10}.  They distinguish between
strict and defeasible inferences and require the syntax of the logic
language of the nodes to contain each instance of a defesaible
inference to allow attacks on inferences.  Apart from this form of
attack they also have attacks on premises and conclusions via a
unidirectional relation specifying that one formula is contrary to
another. ASPIC builds on an extended notion of \emph{abstract
  argumentation system}~\cite{Vreeswijk_1997} and was generalised to
\aspic \cite{Prakken_2010} by partitioning  inferences and facts
into an undeniable and a defeasible part \cite{Modgil_2013}. %
\section{The \acl{T-AIF} T-AIF}
\label{sec:taif}

Aristotle~\cite{2006rhetoric} noted that argument content is not the only relevant aspect for human argument evaluation.
He distinguished between three means of persuasion forming the so called \emph{Aristotelian Trichotomy}:
\begin{description}
  \item[Logos:] \textit{Appeal to logic}. The structure of the argumentation both of individual reasoning steps and the overall interaction of multiple arguments, e.g. argumentation schemes used.
  \item[Ethos:] \textit{Appeal to authority}. The properties of the speaker relevant for the evaluation of argument, e.g. credibility or moral values.
  \item[Pathos:] \textit{Appeal to emotion}. The emotional aspects of an argument, e.g. enthusiasm of the speaker or intended emotional reaction from the audience.
\end{description}
Surprisingly, to our knowledge all formal representations for
argumentation are mostly focused on the logos aspect of argumentation.
The \ac{T-AIF} aims to incorporate all three aspects into the formal
representation to allow representing arguments in a richer way.  It
consists of three interconnected parts representing different aspects
of the argumentation.

\subsection{Trust Network}
\label{subsec:trust}
Building on ideas of Lawrence et al.~\cite{Lawrence_2017}, we include the speakers into the argumentation representation.
Other entities such as politicians, news organisations, or other groups that might participate in the discussion through citation or quotation are also included.
We call these \emph{E-nodes} given that they represent \textbf{E}ntities as unifying concept over active and passive participants of the conversation.

While Lawrence et al.\ use their speaker nodes mostly for statistics
and grouping, we also use them to represent ethos and pathos through
weighted edges in the graph.  The ethos aspect is incorporated by
weighted edges between E-nodes capturing the notion of trust of active
participants towards other entities forming a trust
network~\cite{DBLP:conf/atal/ParsonsTSMC11}.  One could have multiple
different trust relations between the entities to capture the
different means of trust described in the literature
\cite{Castelfranchi1998,Demolombe_2004,Herzig_2009,Parsons_2014,Rodenhauser2014Trust}
but in the following we are mostly concerned with trust as the
confidence an actor has in another actor's utterances.

The grade of commitment of the speakers towards their illocutions is a pathos aspect represented in the graph.
Distinguishing between different levels of commitment or tracking the commitment of actors is mostly studied in the context of dialogue games \cite{WaltonKrabbe95,mcburney2001agent,mcburney2002games,Wells_2012,DBLP:journals/argcom/BudzynskaJRS16} but we believe it has relevant interconnections with a speaker's ethos.
Strongly committing to propositions obligates an actor to defend them when challenged and should have a negative effect on her trustworthiness when this fails \cite{Ogunniye_2018}. %
In contrast, an actor voicing arguments for the opposing view besides her own should receive increased trust due to seeming well-informed and not being afraid of attacks on her own stance.

\subsection{Dialogue Structure}
\label{subsec:dialogue}
This part of the argumentation representation captures the raw locutions the actors have put forth.
The \textbf{L}ocutions are represented as textual labels in \emph{L-nodes}.
Connecting the L-nodes via the reply relation forms a lattice structure.
Intuitively one would assume this to form a tree structure but an utterance may reference multiple other locutions resulting in multiple parents.

Apart from the reply relation, locutions are connected to moves from a dialogue game giving meaning to their interconnections.
Moreover, they are connected to illocution nodes (described in Section~\ref{subsec:argument}) via illocutionary force edges as seen widely in previous work~\cite{DBLP:conf/comma/ReedWBD10,budzynska2011speech,DBLP:conf/comma/BudzynskaJKRSSY14,DBLP:journals/argcom/BudzynskaJRS16}.
Note that in the interest of clarity, our representation uses typed labelled edges to represent illocutionary force and dialog relations in favour of having YA- and TA-nodes.

It is beneficial to keep the locutions of the dialogue in the
representation of the argumentation to enable a broader applicability
of the format as a source structure containing both argumentation and
dialogue.  This structure serves as witness for the temporal
development of the argumentation, and thus e.g.\ allows penalizing
unfavourable moves or decisions in the dialogue game as proposed by
Walton~\cite{Walton_84}.  Penalizing or rewarding based on logical
content as briefly mentioned at the end of Section~\ref{subsec:trust}
also becomes expressible as a result of having the locutions as nodes
in T-AIF graphs.

\subsection{Argument Map}
\label{subsec:argument}

Due to unknown background knowledge of the discussion participants and
enthymemes dominating real argumentative discussions, we cannot assume
the represented argument maps to be complete without unknown parts.
From the dialogue relations we can infer argumentative relations even
if the precise reasoning patterns in use are not clear.  We assume
that the participants of the discussion voice complaints if the
connections between propositions are unclear to a human participant
with relevant background knowledge.  Hence, we assume that enthymemes
and skipped reasoning steps are not fallacious but covered by
knowledge available at the time.

Our argument map is similar to the one defined in the \ac{AIF}
specification~\cite{CHES_EVAR_2006} but our \emph{I-nodes} are
labelled with logical formulas (without committing to a particular
logic) representing the propositional content of the
\textbf{I}llocution.  Attacks and supports have schemes similar to the
ones proposed by the \ac{AIF} but restrict the shape of participating
propositions according to their nature (e.g.\ as specified by
Walton~\cite{Walton95} or Parsons et al.~\cite{Parsons_2014}).  The
\textbf{A}pplication of these schemes is represented in the graph by
\emph{SA-nodes} and \emph{AA-nodes} for \textbf{S}upport and
\textbf{A}ttacks, respectively.  Exceptions to these schemes are
represented as propositions of specified shape connecting to the
application nodes directly.  We allow E-nodes to be connected via
attack and support nodes to express ethical requirements on entities or
infer ethical properties of these respectively.

We leave out the preference nodes from \ac{AIF} in our formalism as preferences are inherently a per entity concept and our argument map represents all actors' contributions.
Preferences of the actors are not inexpressible this way but are rather expressed through trust and commitment in the sense that an actor likely prefers her own arguments or those made by highly truted entities over those made by untrusted entities.

Due to elided reasoning steps the uttered propositions in real argumentative conversations might not have the required shapes for the reasoning patterns they are involved in (see \autoref{ex:elided}).
Filling these reasoning steps from common sense and common knowledge is an easy task for humans (albeit not uniquely) but a very hard task for machines as analysed by Boltužić and Šnajder~\cite{Boltuzic:Snajder:16}.
\begin{example}[Elided Steps in Typical Reasoning]\label{ex:elided}
  \textquote{Experts say that Brexit would hurt the economy, so we should vote against Brexit.}
  Here the main argumentation scheme involved is \emph{Argument from Expert Opinion} based on the fact that experts are quoted.
  The drawn conclusion by the definition of that scheme~\cite{walton_reed_macagno_2008} would be that \textquote{Brexit would hurt the economy} and not that \textquote{UK citizens should vote against Brexit}.
\end{example}
We resort to accepting this lack of information and capturing the unknown reasoning patterns by having default inferences connecting the perceived to the required proposition.
So in the case of \autoref{ex:elided} we would have the conclusion of the argument from expert opinion be that \textquote{Brexit would hurt the economy} and add a default inference from there to \textquote{UK citizens should vote against Brexit}.
Similarly, we add required premises and exceptions to the scheme applications to complete enthymemes without committing entities to these.

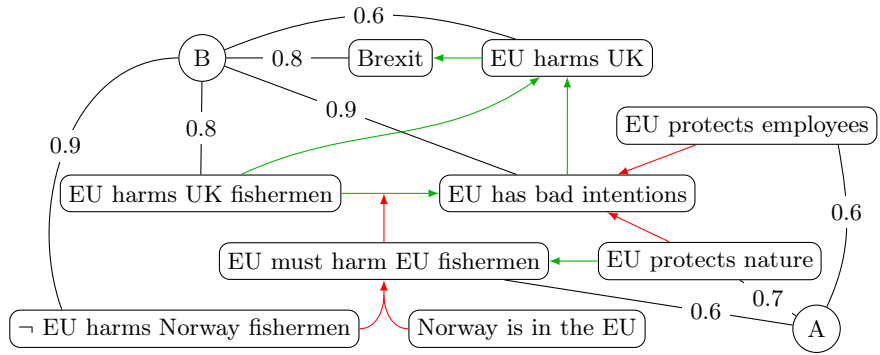
\begin{figure}[ht]
  \centering
  \begin{tikzpicture}[->, -latex, node distance = 4em, supp/.style={green!70!black}, att/.style={red}, prop/.style={draw, rectangle, rounded corners, minimum height = 1.5em}, act/.style={draw, circle}, link/.style={-}]
    \node [prop](Brexit) {Brexit};
    \node [prop, right = 2em of Brexit](EUbadUK) {EU harms UK};
    \node [prop, below = 4em of EUbadUK](EUbad) {EU has bad intentions};
    \node [prop, left = of EUbad](UKfish) {EU harms UK fishermen};
    \node [prop, below = 2em of $(UKfish)!0.5!(EUbad)$](Mfish) {EU must harm EU fishermen};
    \coordinate [below = 2em of Mfish] (mp) {};
    \node [prop, left = 1em of mp](NWfish) {$\neg$ EU harms Norway fishermen};
    \node [prop, right = 1em of mp](NWEU) {Norway is in the EU};
    \node [prop, right = 2em of $(EUbad)!0.5!(EUbadUK)$](EUprotect) {EU protects employees};
    \node [prop, right = 2em of Mfish](EUnature) {EU protects nature};

    \node [act, left = 5em of Brexit](B) {B};

    \draw [link] (B) to node [fill=white] {0.8} (Brexit);
    \draw [link, bend left = 20] (B) to node [fill=white] {0.6} (EUbadUK);
    \draw [link] (B) to node [fill=white, pos = 0.4] {0.9} (EUbad);
    \draw [link] (B) to node [fill=white] {0.8} (UKfish);
    \draw [link, in = 110, out = 180] (B) to node [fill=white] {0.9} ($(NWfish.north west)!0.3!(NWfish.north)$);

    \node [act, right = 6em of NWEU](A) {A};

    \draw [link] (A) to node [fill=white, pos = 0.3] {0.6} (Mfish);
    \draw [link] (A) to node [fill=white] {0.7} (EUnature);
    \draw [link, out = 65, in = -80] (A) to node [fill=white, pos = 0.6] {0.6} ($(EUprotect.south east)!0.3!(EUprotect.south)$);

    \draw [supp] (EUbadUK) to  (Brexit);
    \draw [supp] (EUbad) to (EUbadUK);
    \draw [supp] (UKfish) to (EUbad);
    \draw [supp, out = 25, in = 215] (UKfish) to (EUbadUK);
    \draw [att] (Mfish) to ($(UKfish)!0.5!(EUbad)$);
    \draw [att, -, in = -90, out = 0] (NWfish) to ($(mp)!0.5!(Mfish)$);
    \draw [att, -, in = -90, out = 180] (NWEU) to ($(mp)!0.5!(Mfish)$);
    \draw [att] ($(mp)!0.5!(Mfish)$) to (Mfish);
    \draw [att] (EUprotect) to (EUbad);
    \draw [supp] (EUnature) to (Mfish);
    \draw [att] (EUnature) to (EUbad);
  \end{tikzpicture}
  \caption{Simplified extract from a Twitter discussion between A and B on Brexit. Red arrows represent attack while green arrows represent support. The weighted black edges represent the commitment of the actors to the propositions.}\label{fig:exarg}
\end{figure}

\section{Semantics}
\label{sec:semantics}

For reasoning about \ac{T-AIF} we define a reduced representing structure and define properties on it as fuzzy formulas to give an example of what is possible within this formalism.
We use the standard \L{}ukasiewicz semantics of fuzzy logic (see~Lukasiewicz and Straccia~\cite{LukasiewiczStraccia08} for an overview):
\begin{definition}[\L{}ukasiewicz Semantics \cite{LukasiewiczStraccia08}]\label{def:luk_sem}
  \begin{align*}
    \mathcal{I}(\phi \land \psi) &= \max(\mathcal{I}(\phi) + \mathcal{I}(\psi) - 1, 0) &
    \mathcal{I}(\phi \lor \psi) &= \min(\mathcal{I}(\phi) + \mathcal{I}(\psi), 1)\\
    \mathcal{I}(\phi \implies \psi) &= \min(1 - \mathcal{I}(\phi) + \mathcal{I}(\psi), 1) &
    \mathcal{I}(\neg \phi) &= 1 - \mathcal{I}(\phi)
  \end{align*}
\end{definition}
In particular, $\phi \iff \psi$ is interpreted into
$1 - |\mathcal{I}(\phi) - \mathcal{I}(\psi)|$ in \L{}ukasiewicz Logic,
which matches the intuition we intend in our formulas.  All sets
mentioned in the following definitions are assumed to be finite unless
mentioned otherwise.
\begin{definition}[\ac{T-AF}]\label{def:taif}
  A \emph{\acl{T-AF}}  $(P_i, P_e, S, \as, \es, (\si_s, I_s)_{s \in S}, (O_x)_{x \in P_e})$ is a structure over sets
  \begin{itemize}
    \item $P = P_i \uplus P_e$ of \emph{propositions} formed from \emph{illocutions} $P_i$ and \emph{entities} $P_e$,
    \item $S = S_{att} \uplus S_{sup}$ of \emph{atomic argumentation schemes} formed from \emph{attack schemes} $S_{att}$ and \emph{support schemes} $S_{sup}$,
  \end{itemize}
  with additional data
  \begin{itemize}
    \item $\as : S \to \N$ representing the \emph{arity} of schemes,
    \item $\es : S \to \N$ representing the \emph{exception arity} of schemes,
    \item $\si_s : \truth^{\as(s)} \to \truth^{\es(s)} \to \truth \to \truth$ representing the \emph{interpretation} of atomic schemes,
    \item $I_s \subseteq P^{\as(s)} \times P^{\es(s)} \times P$
      representing \emph{atomic \aap}, relating the \emph{givens} or
      \emph{premises} and the \emph{exceptions} to the \emph{claim} or
      \emph{conclusion}, and
    \item $O_x : P \to (\truth \to \truth)$ representing \emph{belief}.
  \end{itemize}
\end{definition}
\begin{remark}[Entities as Propositions]\label{rem:ent_prop}
  Argumentation actions can include entities as premises and
  conclusions.  An entity seen as a proposition is to be read as the
  trustworthiness of the entity.  Therefore, an \emph{Ad Hominem}
  argument against entity~$x$ would be an \aas with~$e$ as conclusion
  prescribing low trustworthiness.  Trust establishment or propagation
  schemes as described by Parsons et al.~\cite{Parsons_2014} can be modelled
  as inferences perscribing high trustworthiness to an entity.
  Additionally, treating entities as propositions enables the trust
  relations decribed in Section~\ref{subsec:trust} to be incorporated into
  the belief predicate $O$.
\end{remark}
\begin{example}[Kinds of inferences in \acp{T-AF}]
  Note that the signature of the scheme interpretation function $\si$ has as additional argument the acceptability of the conclusion.
  It can therefore be seen as a fuzzy predicate over the combination of givens, exceptions, and claim where \emph{givens} and \emph{exceptions} are the propositions related to a \emph{claim} by an \aas.
  This allows expressing different concepts, and combinations thereof, via this single
  notion:
  \begin{itemize}
    \item \emph{Exceptions} can produce a predicate that accepts all truth values of the claim as the exception defeats the \aas;
    \item \emph{Weighted premises} as seen in the specification of Walton-style schemes can be incorporated through linear combination of the givens;
    \item \emph{Necessary conditions} can produce a predicate
      specifying that the claim can not be accepted unless the givens
      are accepted: $\neg p \implies \neg c$ (read, e.g., as a fuzzy
      formula). This is what Boudhar et
      al.~\cite{DBLP:conf/icaart/BoudharNR12} call \emph{necessary
        support};
    \item \emph{Sufficient conditions} can produce a predicate specifying that the claim has to be accepted when the givens are accepted: $p \implies c$. This is what Boella et al.~\cite{DBLP:conf/comma/BoellaGTV10} call \emph{deductive support};
    \item \emph{Inhibiting conditions} can produce a predicate specifying that the claim can not be accepted unless the givens are rejected: $p \implies \neg c$;
    \item \emph{Disjunctive conditions} can produce a predicate specifying that the givens or the claim need to be accepted: $\neg p \implies c$;
  \end{itemize}
\end{example}
Similarly, belief is interpreted as a predicate $O$ rather than a
single truth value, thus capturing belief and disbelief as well as
uncertainty.
\begin{definition}[Scheme Interpretation]\label{def:scheme_inter}
  The scheme interpretation is required to adhere to the given type of
  argumentation scheme: An attack (support) is interpreted as a fuzzy
  predicate that allows increasing (decreasing) acceptance of the
  claim if the acceptance of the premises increases.  This is made
  explicit in the following constraints for scheme interpretation:
  \begin{align*}
    \bigwedge_{s \in S_{sup}; \bar{p}, \bar{q} \in \truth^{\as(s)}; c \in \truth} \bar{p} \leq \bar{q}\implies \bigvee_{c' \in \truth} c \leq c' \land \si_s(\bar{p}, \bot^{\es(s)}, c) \leq \si_s(\bar{q}, \bot^{\es(s)}, c')\\
    \bigwedge_{s \in S_{att}; \bar{p}, \bar{q} \in \truth^{\as(s)}; c \in \truth} \bar{p} \leq \bar{q} \implies \bigvee_{c' \in \truth} c' \leq c \land \si_s(\bar{p}, \bot^{\es(s)}, c) \leq \si_s(\bar{q}, \bot^{\es(s)}, c')
  \end{align*}
\end{definition}
\begin{example}[Logically: Argument from Position to Know]\label{ex:log_pos}
  Consider a logical representation of \emph{Argument from Position to
    Know} (\autoref{ex:pos}) using an epistemic modality $B_a$, read
  ``$a$ believes that'', and a modality~$L_a$ for locution, read ``$a$
  said that'':
  \[\prftree[r]{(Position to Know)}{B_a A \iff A}{L_a A}{L_a A \implies B_a A}{A}\]
  Critical questions 1 and 3 are modelled as attacks against truth of the premises while question 2 would be an attack against an added honesty premise $L_a A \implies B_a A$.
  As E-nodes are treated as propositions for their honesty the third premise would be represented simply by $a$.
  Note that the inference is deductively valid and therefore can have no exceptions but only attacks on premises.
  This also leads to a uniform weighting of premises;  hence the interpretation for the scheme would correspond to the formula $\Big((B_a A \iff A) \land L_a A \land a\Big) \implies A$ involving illocutions $B_aA \iff A$, $L_aA$, and~$A$ as well as the entity~$a$.
\end{example}
Due to having diverse and joint notions of inference, forming complex attacks to derive sensible semantics becomes more challenging~\cite{Amgoud_2017,Cohen_2018}.
\begin{definition}[Composite Schemes]\label{def:scheme_comp}
  \newcommand*{\comp}[3]{#1;_{#2}#3}
  Define the extended sets of \emph{complex support (attack) schemes} $S_{sup}'$ ($S_{att}'$) and their union $S'$ of \emph{complex schemes} inductively:
  \begin{itemize}
    \item $S_{sup} \subseteq S_{sup}'$ and $S_{att} \subseteq S_{att}'$
    \item Given schemes $s \in S_{sup}', t \in S'$ and $i \in \as(t)$ define the \emph{composite scheme} $\comp sit$.
  Depending on wether $t$ was an attack or a support, $\comp sit$ belongs to  $S_{att}'$ or $S_{sup}'$, respectively.
  \end{itemize}
  The composite scheme $s;_it$ is interpreted as follows:
  \begin{align*}
    &\as(\comp sit)  = \as(s) + \as(t) - 1 \\
    &\es(\comp sit)  = \es(s) + \es(t) \\
  &I_{\comp sit} = \{(u_1, \dots, u_{i - 1}, \bar{v}, u_{i + 1}, \dots, u_{\as(t)}, \bar{e}, \bar{f}, p) \mid (\bar{v}, \bar{e}, \pi_i\bar{u}) \in I_s, (\bar{u}, \bar{f}, p) \in I_t\}\\
    &\si_{\comp sit}(u_1, \dots, u_{i-1}, \bar{v}, u_{i+1}, \dots, u_{\as(t)}, \bar{e}, \bar{f}, p) =\\ &\qquad \textstyle\bigvee_{b \in \truth} \si_s(\bar{v}, \bar{e}, b) \land \si_t(u_1, \dots, b, \dots, u_{\as(t)}, \bar{f}, p)
  \end{align*}
  Note that $S_{att}'$ or $S_{sup}'$ may be infinite if there are
  cycles in the graph that allow infinite composition.
\end{definition}

\noindent For classifying extensions in our T-AFs, we follow the
approach of graded bipolar frameworks, which define a fuzzy labelling
specifying the acceptability of an argument.  Using graded acceptance,
uncertainty and preferences can be incorporated.
\begin{definition}\label{def:labelling}
  A \emph{labelling} for a \ac{T-AF} is a function~$l : P \to \truth$
  assigning to each proposition a degree of acceptance.  Let $L$ denote
  the set of all labellings.
\end{definition}
Properties of labellings and actors can then be expressed
as fuzzy formulas over \acp{T-AF}.
Entities can be grouped based on the similarity of their beliefs, e.g.\ in order to reason about group dynamics.
\begin{definition}[Similarity]\label{def:similar}
  Two actors~$x, y$ in a \ac{T-AF} are considered \emph{similar} if they have voiced similar belief.
  $$\Sim(x, y) \defeq \textstyle\bigwedge_{l \in L; p \in P} O_x(p, l(p)) \iff O_y(p, l(p))$$\end{definition}
\begin{definition}[Attack \& Support]\label{def:att_supp}
  A proposition~$p$ is \emph{attacked} under a given labelling~$l$ if there is an attack (complex or atomic) with accepted givens, rejected exceptions, and claim $p$:
  $$\textstyle\Att_l(p) \defeq \bigvee_{s \in S_{att}'; (\bar{q}, p) \in I_s}\bigwedge_{i \in \as(s)} l(\pi_i\bar{q}) \land \bigwedge_{i \in \es(s)} \neg l(\pi_{\as(s) + i}\bar{q})$$
  The definition of \emph{support} $\Sup_l(p)$ is similar but with $s \in S_{sup}'$.
\end{definition}
In Dung-style \acp{AAF} (\autoref{def:dung_aaf}), the semantics is
mostly defined in terms of a notion of \emph{defense}, which needs to
be suitable adapted in the presence of supports and exceptions:
\begin{definition}[Defense]\label{def:defense}
  A proposition~$p$ is \emph{defended} under a labelling~$l$ if every
  attack against $p$ is \emph{defeated}.  That is, either a given of
  the attack is attacked or an exception is supported by accepted
  propositions:
  $$\textstyle\Def_l(p) \defeq \bigwedge_{s \in S_{att}'; (\bar{q}, p) \in I_s} \bigvee_{i \in \as(s)}\Att_l(\pi_i\bar{q}) \lor \bigvee_{i \in \es(s)} \Sup_l(\pi_{\as(s) + i}\bar{q})$$
\end{definition}
In \acp{AAF}, the notion of attack induces a notion of
\emph{conflict-freeness}.  For \acp{T-AF}, we generalize this concept
to a notion of \emph{consistency} effectively incorporating both
attack and support.
\begin{definition}[Consistency]\label{def:consistent}
  A labelling~$l$ for a \ac{T-AF} is \emph{consistent} if the assigned labels are in accordance with the \aap of the system.
  $$\textstyle\Co(l) \defeq \bigwedge_{s \in S; (\bar{v}, p) \in I_s} \si_s(l[\bar{v}], l(p))$$
\end{definition}
\noindent (To see that this does indeed generalize the standard notion
of conflict-freeness, introduce a unary attack scheme and equip it
with mutual exclusion between the given and the claim as the schema
interpretation.)
\begin{remark}
  Note that $\Co(l)$ can be seen as the evaluation of the formula $\bigwedge_{s \in S; (\bar{v}, p) \in I_s} s(\bar{v}, p)$ under the labelling~$l$ but we prefer to write it as defined because we later need to talk about different labellings in the same formula.
\end{remark}
The \emph{extensions} in the style of Dung~\cite{Dung_1995} become
fuzzy predicates defined in terms of our definition of defense and
consistency --- that is, a given labelling will not be absolutely stable,
preferred etc., but rather have these properties to a certain degree,
in accordance with our general consideration of weights. We emphasize
that this applies in particular also to preferred and grounded
labellings, which are now `maximal' or `least' not in an absolute
but in a fuzzy sense.
\begin{definition}[Admissible Labelling]\label{def:adm}
  A labelling~$l$ is \emph{admissible} if it is consistent and each accepted proposition is defended in $l$:
  $$\textstyle\Al(l) \defeq \Co(l) \land \bigwedge_{p \in P} l(p) \implies Def_l(p)$$
\end{definition}
\begin{definition}[Stable Labelling]\label{def:stable}
  A labelling~$l$ is \emph{stable} if it is consistent and each rejected proposition is attacked under $l$:
  $$\textstyle\Sl(l) \defeq \Co(l) \land \bigwedge_{p \in P} \neg l(p) \implies \Att_l(p)$$
\end{definition}
\begin{definition}[Preferred Labelling]\label{def:pref}
  A labelling is \emph{preferred} if it is a maximal admissible labelling:
  $$\Pl(l) \defeq \Al(l) \land \bigwedge_{l' \in L} \bigg(\Al(l') \implies \Big(\bigwedge_{p \in P}l(p) \implies l'(p)\Big) \implies \bigwedge_{p \in P}l'(p) \implies l(p)\bigg)$$
\end{definition}
\begin{definition}[Complete Labelling]\label{def:comp}
  A labelling is \emph{complete} if it is admissible and accepts each defended proposition:
  $$\textstyle\Cl(l) \defeq \Al(l) \land \bigwedge_{p \in P}\Def_l(p) \implies l(p)$$
  \draftonly{Dung: An admissible set $S$ of arguments is called a complete extension iff each argument, which is acceptable w.r.t. $S$, belongs to $S$.}
\end{definition}
\begin{definition}[Grounded Labelling]\label{def:gro}
  A labelling is \emph{grounded} if it is a least complete labelling of the \ac{T-AF}:
  $$\textstyle\Gl(l) \defeq \Cl(l) \land \bigwedge_{l' \in L} \Cl(l') \implies \bigwedge_{p \in P}l(p) \implies l'(p)$$
  \draftonly{Dung: The grounded extension of an argumentation framework $AF$ denoted by $GE_{AF}$, is the least fixed point of $F_{AF}$.}
\end{definition}
Apart from translating the definitions of Dung \acp{AAF}~\cite{Dung_1995} we can define exemplary properties of the actors in the dialogue specific to our extended setting.
\begin{definition}[Agreement]\label{def:agreement}
  A labelling~$l$ \emph{agrees} with an actor~$x$ if~$x$'s beliefs are retained in $l$:
  $$\textstyle\Ag(l, x) \defeq \bigwedge_{p \in P} O_x(p, l(p))$$
\end{definition}
Having defined agreement opens possibilities to express interesting properties like rationality of an actor.
\begin{definition}[Rationality]\label{def:rationality}
  An actor~$x$ is \emph{rational} if her belief can be extended to as a consistent position:
  $$\textstyle\Ra(x) \defeq \bigvee_{l \in L}\Co(l) \land \Ag(l, x)$$
\end{definition}
\begin{definition}[Justified Trust]\label{def:just_trust}
  An actor~$x$ has \emph{justified trust} if the belief of actors $x$ trusts can be consistent in the same labelling:
  $$\textstyle\Jt(x) \defeq \bigvee_{l \in L}\Co(l) \land \bigwedge_{y \in P_e}\Big((O_x(y, l(y)) \land l(y)) \implies \Ag(l, y)\Big)$$\mgnote{``Aktoren denen x glaubt'' ist jetzt etwas unschön finde ich.}
\end{definition}
An actor~$x$ having justified trust not only expresses that the actors $x$ trusts have rational beliefs but also that these beliefs are consistent with eachother.
\begin{definition}[Trust Compliance]\label{def:trust_comp}
  A labelling~$l$ is \emph{trust compliant} for an actor~$x$ if belief of actors trusted by~$x$ is reflected in~$l$:
  $$\textstyle\Tc(l, x) \defeq \bigwedge_{p \in P_i} \Big(\bigvee_{y \in P_e} O_x(y, l(y)) \land l(y) \land O_y(p, l(p))\Big)$$
\end{definition}
Trust compliance expresses that $x$ trusts her trusted actors for validity in the sense of Demolombe~\cite{Demolombe_2004} and reflects that in her beliefs (the labelling~$l$).
Combining all trusted actors via disjunction captures a similar notion to parallel path composition as defined by Parsons et al.~\cite{DBLP:conf/atal/ParsonsTSMC11}.

\section{\aspic Translation}
\label{sec:aspic_trans}
Given that the argument map in \ac{T-AIF} is very similar to \ac{AIF} and hence can be easily converted, we obtain \aspic semantics using an existing translation~\cite{DBLP:conf/comma/BexPR10}.
That translation can be improved with the additional information available in our extended formalism to obtain per-actor semantics from the translation.

For constructing the argumentation theory we modify the translation
\cite{DBLP:conf/comma/BexPR10} by forming a separate knowledge base
for each actor.  An ordering on the formulas in the knowledge base can
be derived from the trust network via the accumulated trust and
commitment of each author.  This semantically results in different
systems for each participant that reflect her own beliefs in the
propositions as well as her trust in the other entities participating
in the discussion.

\section{Conclusion and future research}
\label{sec:conclusion_future}
We have proposed the \emph{\acl{T-AIF}}, aimed at representing
argumentation resulting from dialogue between multiple actors,
capturing aspects from all three areas of the aristotelian trichotomy.
Our format is inspired by the \ac{AIF} and especially its extensions \aifp and
\ac{S-AIF}. The ethos aspect is captured by relating entities via
trust and allowing their ethical properties to be involved in
arguments.  Relating entities to their illocutions incorporates the
strength of their commitment as an aspect from the domain of pathos.
Finally, the argument map represents the logos aspect of argumentation,
capturing the logical connections between the propositions.  Given
that the \ac{AIF} and the existing extending formalisms are specified
using the \ac{OWL}, we plan to formalize our format in an ontology as
well.

Our main contribution in this paper is to provide a formalism for
representing and reasoning about more than just the logos aspect of
argumentation.  This enables a very natural treatment, e.g., of Ad
Hominem arguments, which affect the acceptance of all arguments made
by the attacked actor.  Having the dialogue
game~\cite{Modgil,DBLP:conf/comma/ReedWDR08} in the representation can
interact nicely with ethical aspects of argumentation by e.g.\
penalizing trustworthiness when illogical moves are made or
undefendable stances are taken.  The representation is also a good
basis for deriving a per-participant semantics incorporating voiced
beliefs and trust relations of actors to approximate what they are
likely to believe.

We have proposed \acp{T-AF} as an exemplary simplified version of
\ac{T-AIF}, aimed at defining properties profiling the participating
actors.  \ac{T-AF}s do not include the dialogue part of \ac{T-AIF},
and there are meta-arguments --- that is, arguments talking about the
discussion itself, which in fact occur rather frequently --- that
become expressible when the dialogue is included in the structure.  An
example of this kind are arguments against an actor's credibility on
the premise that she has retracted previous statements she had high
commitment to when counterarguments were given.

Since the tooling around the \ac{AIF} ecosystem only supports some
features of \ac{T-AIF}, we plan to provide tool support for creating,
storing, and visualizing argument structures in our format.
The envisioned tool support will also include reasoning capabilities
beyond the described \aspic translation for working on a given
\ac{T-AIF} graph directly.

As maybe apparent from our choice of examples, we plan to evaluate our
formalism on a dataset of argumentative discourse on Twitter,
specifically concerning Brexit (pre-referendum).  Ultimately, we
intend to construct \ac{T-AIF} representations automatically from
written discussions using a structured argumentation mining algorithm.

\bibliographystyle{myabbrv}
\bibliography{literature.bib}

\clearpage
\appendix
\section{Appendix}
\label{sec:appendix}

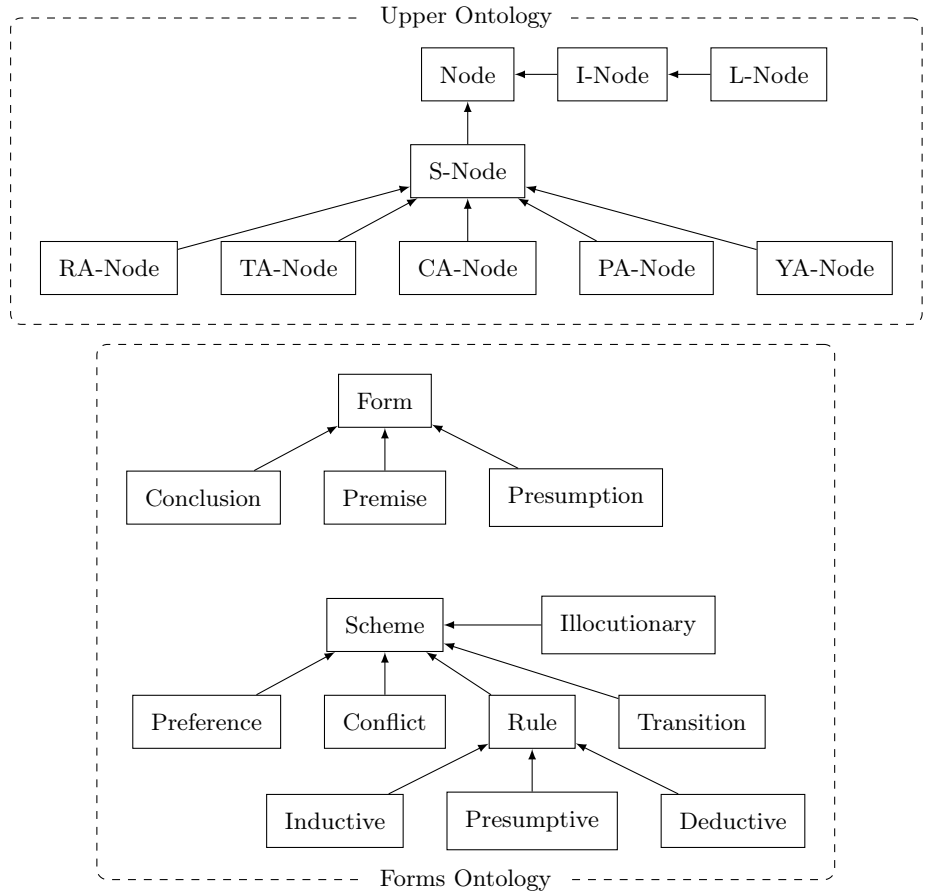
\begin{figure}[ht]
  \centering
  \begin{tikzpicture}[every node/.style={draw, rectangle, inner sep = .75em}, ->, -latex, node distance = 1.75em]
    \node (node) {Node};
    \node [below = of node] (S) {S-Node};
    \node [right = of node] (I) {I-Node};
    \node [right = of I] (L) {L-Node};
    \node [below = of S] (CA) {CA-Node};
    \node [left = of CA] (TA) {TA-Node};
    \node [left = of TA] (RA) {RA-Node};
    \node [right = of CA] (PA) {PA-Node};
    \node [right = of PA] (YA) {YA-Node};

    \draw (S) -- (node);
    \draw (I) -- (node);
    \draw (L) -- (I);
    \draw (CA) -- (S);
    \draw (TA) -- (S);
    \draw (RA) -- (S);
    \draw (PA) -- (S);
    \draw (YA) -- (S);

    \node [fit = (S) (I) (L) (CA) (TA) (RA) (PA) (YA), rounded corners, label={[label distance = -1.2em, fill = white]Upper Ontology}, inner sep = 1.2em, dashed] (upper) {};

    \node [below left = 2em and -17.25em of upper](form) {Form};
    \node [below = of form](prem) {Premise};
    \node [left = of prem](conc) {Conclusion};
    \node [right = of prem](pres) {Presumption};

    \draw (prem) -- (form);
    \draw (conc) -- (form);
    \draw (pres) -- (form);

    \node [below = 3em of prem](scheme) {Scheme};
    \node [below = of scheme](conf) {Conflict};
    \node [left = of conf](pref) {Preference};
    \node [right = of conf](rule) {Rule};
    \node [below = of rule](presr) {Presumptive};
    \node [left = of presr](ind) {Inductive};
    \node [right = of presr](ded) {Deductive};
    \node [right = of rule](tra) {Transition};
    \node [right = 4em of scheme](illo) {Illocutionary};

    \draw (conf) -- (scheme);
    \draw (pref) -- (scheme);
    \draw (rule) -- (scheme);
    \draw (tra) -- (scheme);
    \draw (illo) -- (scheme);
    \draw (presr) -- (rule);
    \draw (ind) -- (rule);
    \draw (ded) -- (rule);

    \node [fit = (form) (prem) (conc) (pres) (scheme) (conf) (pref) (rule) (tra) (illo) (presr) (ind) (ded), rounded corners, label={[label distance = -1.2em, fill = white]below:Forms Ontology}, inner sep = 1.2em, dashed] (lower) {};

  \end{tikzpicture}
  \caption{The \aifp Ontology. The boxes are concepts and the arrows represent the inclusion relation. L-nodes, TA-nodes, and Transition Schemes~\cite{DBLP:conf/comma/ReedWDR08} as well as YA-nodes and Illocutionary Schemes~\cite{DBLP:conf/comma/ReedWBD10} were added to \ac{AIF}.}\label{fig:aif}
\end{figure}

\end{document}